\documentclass[sigplan,screen]{acmart}

\renewcommand\footnotetextcopyrightpermission[1]{} 
\begin{document}

\title{Customizable LLM-Powered Chatbot \\for Behavioral Science Research}
\pagestyle{plain}

\author{Zenon Lamprou} 
\authornote{Both authors contributed equally to this research.}
\orcid{0000-0003-0042-5036}
\email{zenon.lamprou@strath.ac.uk}
 \affiliation{
   \institution{University of Strathclyde}
   \city{Glasgow}
   \country{Scotland}
}
\author{Yashar Moshfeghi}
\authornotemark[1]
\orcid{0000-0003-4186-1088}
\email{yashar.moshfeghi@strath.ac.uk}
 \affiliation{
   \institution{University of Strathclyde}
   \city{Glasgow}
   \country{Scotland}
}

\renewcommand{\shortauthors}{Lamprou and Moshfeghi}

\begin{abstract}
The rapid advancement of Artificial Intelligence has resulted in the advent of Large Language Models (LLMs) with the capacity to produce text that closely resembles human communication. These models have been seamlessly integrated into diverse applications, enabling interactive and responsive communication across multiple platforms. The potential utility of chatbots transcends these traditional applications, particularly in research contexts, wherein they can offer valuable insights and facilitate the design of innovative experiments. In this study, we present a Customizable LLM-Powered Chatbot (CLPC)\footnote{The CLPC system can be downloaded from \url{https://github.com/NeuraSearch/CLPC}.}, a web-based chatbot system designed to assist in behavioral science research. The system is meticulously designed to function as an experimental instrument rather than a conventional chatbot, necessitating users to input a username and experiment code upon access. This setup facilitates precise data cross-referencing, thereby augmenting the integrity and applicability of the data collected for research purposes. It can be easily expanded to accommodate new basic events as needed; and it allows researchers to integrate their own logging events without the necessity of implementing a separate logging mechanism. It is worth noting that our system was built to assist primarily behavioral science research but is not limited to it, it can easily be adapted to assist information retrieval research or interacting with chat bot agents in general.
\end{abstract}



\maketitle

\section{Introduction}
The swift advancement of Artificial Intelligence has resulted in the advent of Large Language Models (LLMs) with the capacity to produce text that closely resembles human communication. These models have been seamlessly integrated into diverse applications, enabling interactive and responsive communication across multiple platforms. A significant implementation of this technology is observed in the development of chatbots, which frequently serve as the initial point of contact in numerous customer service and assistant roles. Nevertheless, the potential utility of chatbots transcends these traditional applications, particularly in research contexts, wherein they can offer valuable insights and facilitate the design of innovative experiments.

In the quest to enhance research methodologies, the CLPC system represents a significant innovation aimed at supporting behavioral scientific investigations. By employing the React framework, CLPC functions efficiently across web and mobile platforms, presenting a responsive user interface that adjusts to various screen dimensions. The system is meticulously designed to function as an experimental instrument rather than a conventional chatbot, necessitating users to input a username and experiment code upon access. This setup facilitates precise data cross-referencing, thereby augmenting the integrity and applicability of the data collected for research purposes.

The principal advantage of CLPC resides in its flexibility and customizability, which are crucial for adaptation to a variety of research scenarios. Users are empowered to modify numerous parameters without necessitating alterations to the foundational code, thereby enabling straightforward experimentation and promoting extensive adoption within the research community. The system accommodates the integration of various LLMs, facilitating a comparative evaluation of model performance across different contexts. By sustaining a robust backend architecture in Python, CLPC permits the effortless incorporation of novel models, thereby enhancing its versatility and applicability in academic and scientific research domains.

\section{Related Work}\label{sec:rel_work}
In every research environment data are the backbone of every statistical analysis and the training of any machine learning model to perform predictions. The field of Information Retrieval (IR) is no different. New data has to be gathered frequently to enrich  current data collections available publicly. To collect data an essential component is a stimuli presentation software . A vital feature of a stimuli presentation software is to have accurate timings for when the stimuli is presented \cite{caton2011event}. Different attempts have been made deploying different techniques and software to achieve the desired latency from using Windows API \cite{ozbeyaz2019stipresoft}, to helping medical professionals in routine testing but also in ICU testing \cite{natus_neuroworks}.CLPC tries to come out of the lab and develop a system that can be used as naturally as possible and bridge the gap of interacting with chatbot Agents in a lab environment oppose to the a more casual environment. 

Furthermore the use of conversational agents have been recently a hot topic in the machine learning community \cite{pinxteren_human-like_2020}. With the introduction of ChatGPT\cite{GPT3Architecture} and Claude\cite{ClaudeAI} just to name a few humans have been interacting with chatbot agents more than ever. Chatbot agents are also utilized extensively now in healthcare \cite{macneill_health_2024} and for therapy \cite{park_survey_2023} to name a few of the domains currently used. Currently these agents operate on their own or 2-3 together. CLPC aims to implement all the most well known chatbot agents and generative LLMs to be interchange and work together on the same conversation.

\section{System architecture}\label{sec:architecture}
\subsection{User Interface}\label{sec:user_interface}
The CLPC System is available for both web and mobile platforms since the front-end is build using the React framework. React offers the flexibility of generating one application that can fit all platforms and can offer a responsive design that can fit almost all screen sizes.

When the user firstly enters the platform is asked to enter a username and the experiment code. Username is used to crosscheck the data logged by the system for the subject. We intent to use CLPC as an experimental tool rather than a normal chatbot thus the experiment code refers to the current experiment so the researches can later on cross check which experiment the subject was performing on the given dataset.

After the user enters their credentials the main chat screen is presented. The user interface itself is pretty simple. A single chat window that the user can write a query and get back the response from the CLPC. On the left hand side you can click the settings icon which currently offers 1 set of parameter customization.
\begin{enumerate}
    \item Change the LLM which will generate the reply.
    \item Change the font size.
    \item Change the line spacing.
\end{enumerate}

On the bottom of the window there is a text box that takes the input from the user and uses the current selected LLM to get back a response. The user can also flag a response as being helpful or not using the thumbs up and thumbs down buttons next to each response.

\begin{figure*}[hbt!]
    \centering
    \includegraphics[width=\linewidth, height=0.21\textheight,keepaspectratio]{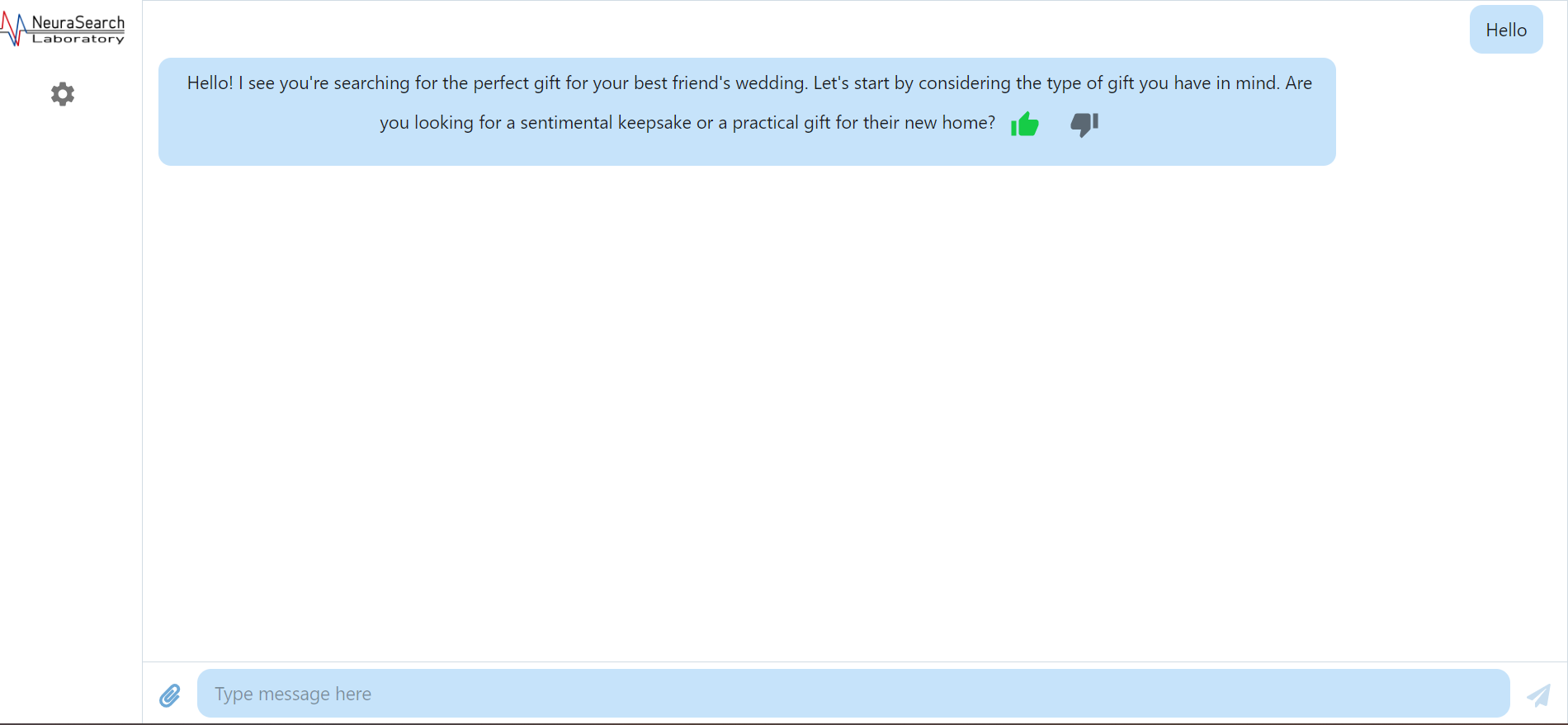}
    \caption{Simple usage of the CLPC. The user sends a message and receives back a response which then proceeds to tag as relevant by pressing the thumbs up button.}
    \label{fig:user_interface}
\end{figure*}
\subsection{Flexibility and Customization}
The CLPC provides significant flexibility and customization for researchers, enabling adaptation to various experimental scenarios with minimal code changes. It can be enhanced with additional functionalities by utilizing the existing system framework, allowing for straightforward implementation of new features. As outlined in Section \ref{sec:user_interface}, the user interface allows  the selection of different LLMs used to generate responses.

CLPC provides a template configuration file that researchers can set prior to experiments to further customize system behavior. Users can add prompts to CLPC to modify the model's responses without requiring code alterations. Furthermore, a second configuration file is available for researchers to input default values before experiments, allowing for comprehensive customization of system parameters and behaviors.
\begin{figure}
    \centering
    \includegraphics[width=\linewidth,keepaspectratio]{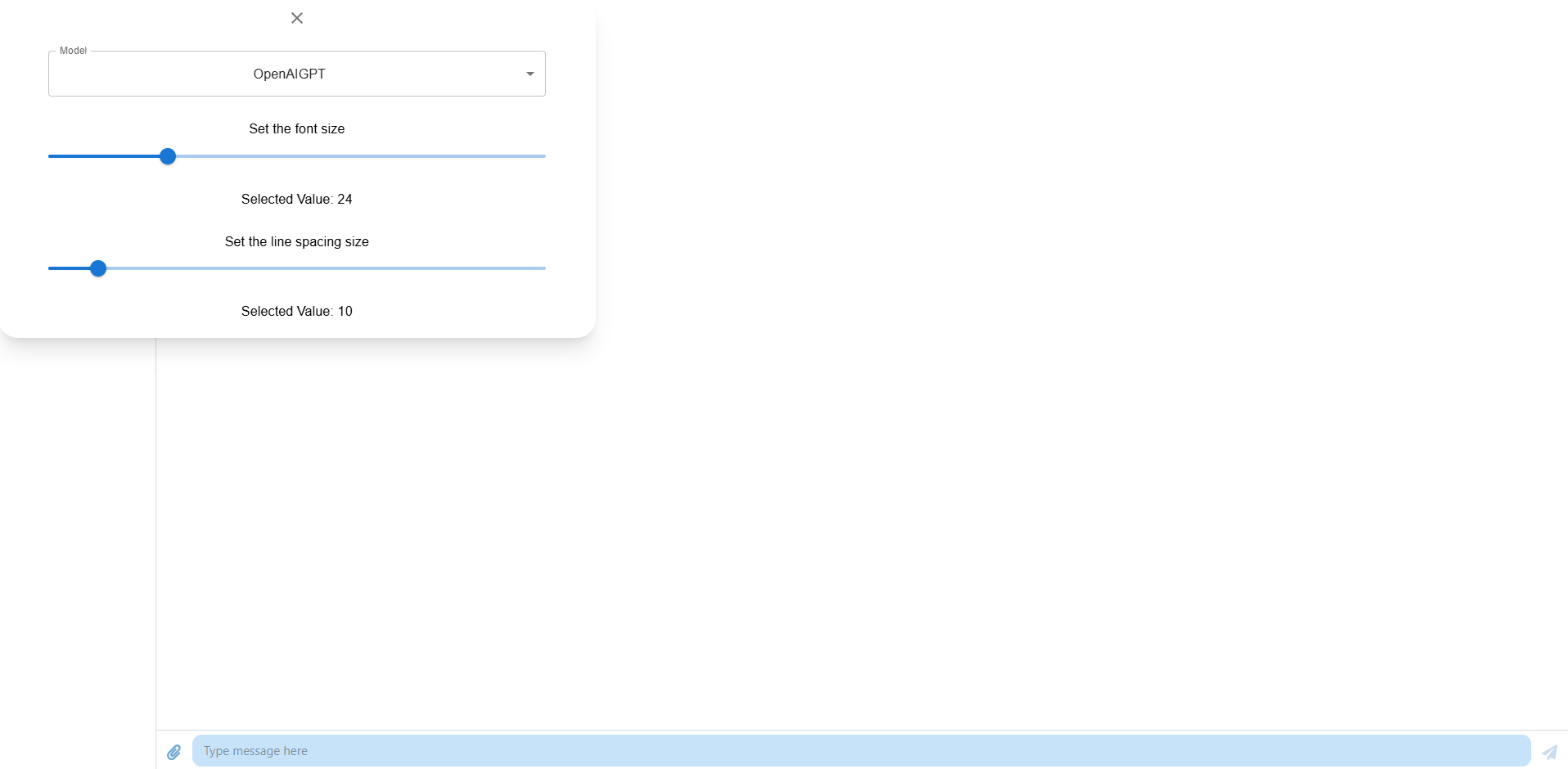}
    \caption{An example of using the configuration provided from the CLPC.}
    \label{fig:settings_screenshots}
\end{figure}
\subsection{Utilizing Different LLMs}
The CLPC system possesses a highly advantageous feature that allows for the incorporation of responses from various LLMs while maintaining the continuity of an ongoing conversation. By tracking the current dialogue at a higher level than the individual LLMs, CLPC enables users to switch between different models on the fly, as illustrated in Figure \ref{fig:settings_screenshots}. This functionality is particularly beneficial in experimental settings, as it allows researchers to evaluate and compare responses from multiple LLMs simultaneously. The entire backend of CLPC is developed in Python, which simplifies the implementation and integration of new models into the system. Additionally, CLPC provides researchers with the flexibility to introduce new or custom models, provided they can be loaded and executed within a Python environment. This design not only enhances the adaptability of CLPC but also fosters a more comprehensive exploration of LLM capabilities in various research contexts.

Furthermore, the CLPC system provides an additional two sets of customization options. Researchers have the ability to modify both the font size and line spacing of the response. The benefit of this customization is twofold. Firstly, it offers a platform that, without any modification to the code, can be utilized by individuals with visual impairments, thereby expanding accessibility to a broader audience. Secondly, it ensures that, when an eye tracker is employed to observe how users read the response, the eye tracking maintains precision in aligning with the user's actual reading behavior, thus preventing any misinterpretation of word placement.
\subsection{Logging Mechanism}\label{sec:log_mechanism}
The CLPC system features a customizable logging mechanism designed to capture various user events and interactions with the interface. This logging capability was developed for two primary purposes: first, to ensure that the system can be easily expanded to accommodate new basic events as needed; and second, to allow researchers to integrate their own logging events with minimal coding effort, without the necessity of implementing a separate logging mechanism. CLPC captures user interactions in both the front-end and back-end, providing a safeguard against data loss in the event of server or UI failures by storing all interactions in a backup location for later retrieval.

By default, the system logs a variety of events, including user actions such as clicking the thumbs-up or thumbs-down buttons, hovering over answer bubbles, sending responses, receiving responses from the language model, and recording the start and end times for displaying responses. The logging mechanism is designed generically, enabling users to implement only the specific events they wish to log while seamlessly integrating these custom events into the existing system. This flexibility enhances the usability of CLPC for researchers conducting experiments and other interactive studies, ensuring comprehensive data collection without unnecessary complexity.

\section{Conclusion and Future Work}\label{sec:conclusion}
In this study, we present CLPC, an innovative web-based chatbot system meticulously designed to enhance behavioral science research initiatives. The defining characteristic of this system is its exceptional flexibility and adaptability, enabling its effective application across diverse research environments while maintaining an intuitive and user-friendly interface. By employing configuration files, we have refined the setup process, thereby obviating the requirement for any coding knowledge or expertise among users. Moreover, CLPC is endowed with a comprehensive logging mechanism that augments the tracking of interactions, supplemented by a versatile trigger system that can be activated promptly or adapted with ease to accommodate additional demands. As we anticipate future developments, our objective is to incorporate an advanced mechanism to effectively manage server delays, thereby extending CLPC's operational capabilities beyond local deployments into wider distributed online settings. Although the system was specifically developed to facilitate behavioral science research, it inherently possesses the potential to be adapted with minimal effort for applications in information retrieval research or for general interactions with chatbot agents across various domains.
\bibliographystyle{ACM-Reference-Format}
 \bibliography{biblio}
\end{document}